\documentclass{article}
\usepackage[utf8]{inputenc}
\usepackage{amsmath, amssymb, graphicx}
\DeclareMathOperator*{\argmin}{arg\,min}
\usepackage{booktabs}
\usepackage{amsthm} 
\usepackage{hyperref}
\usepackage{algorithm}
\usepackage{algpseudocode}
\usepackage[numbers]{natbib}
\usepackage{booktabs} 
\usepackage{array}    
\usepackage{caption}
\usepackage[a4paper, total={6in, 8in}, left=0.5in, right=0.5in, top=1in, bottom=1in]{geometry} 
\usepackage{algorithm}
\usepackage{algorithmicx}
\usepackage{algpseudocode}
\usepackage{color}
\usepackage{tikz}
\usetikzlibrary{positioning, arrows.meta, shapes.geometric}
\usepackage{multirow}

\usepackage[below]{placeins}
\usepackage{sidecap}
\usepackage{graphicx}
\usepackage{moreverb}   
\usepackage{hyperref}
\usepackage{fancyhdr}
\usepackage{wrapfig}
\usepackage[numbers]{natbib}

\numberwithin{equation}{section}

\title{Evaluating generative models for synthetic financial data}
\author{Christophe D. Hounwanou \thanks{christophe.hounwanou@aims.ac.rw} \\ 
        AIMS Rwanda, \\ 
        African Institute for Mathematical Sciences \\
        \and
        Pierre Ntakirutimana
        \thanks{pntakiru@alumni.cmu.edu} \\ 
        College of Engineering,\\
        Carnegie Mellon University \\
        \and
        Ya\'e Ulrich Gaba\thanks{yaeulrich.gaba@gmail.com} \\ 
        Sefako Makgatho Health Sciences University (SMU)\\ Pretoria, South Africa \\
        \&\\
         AI Research and Innovation Nexus for Africa (AIRINA Labs) \\ AI.Technipreneurs, B\'enin
}

\begin{document}

\maketitle
\begin{abstract}
Data scarcity and confidentiality in finance often impede model development and robust testing. This paper presents a unified multi-criteria evaluation framework for synthetic financial data and applies it to three representative generative paradigms: the statistical ARIMA–GARCH baseline, Variational Autoencoders (VAEs), and Time-series Generative Adversarial Networks (TimeGAN). Using historical S\&P 500 daily data, we evaluate fidelity (Maximum Mean Discrepancy, MMD), temporal structure (autocorrelation and volatility clustering), and practical utility in downstream tasks, specifically mean–variance portfolio optimization and volatility forecasting. Empirical results indicate that ARIMA–GARCH captures linear trends and conditional volatility but fails to reproduce nonlinear dynamics; VAEs produce smooth trajectories that underestimate extreme events; and TimeGAN achieves the best trade-off between realism and temporal coherence (e.g., TimeGAN attained the lowest MMD: $1.84\times10^{-3}$, average over 5 seeds). Finally, we articulate practical guidelines for selecting generative models according to application needs and computational constraints. Our unified evaluation protocol and reproducible codebase aim to standardize benchmarking in synthetic financial data research.

\end{abstract}

\section{Introduction}

Financial modeling and quantitative research depend critically on access to large volumes of high-quality data for forecasting, risk assessment, and portfolio optimization. Yet, financial datasets are often constrained by strict privacy regulations, proprietary restrictions, and significant acquisition costs, which hinder open access and reproducibility \cite{bollerslev1986generalized, patki2016synthetic, xu2019modeling}. This data scarcity limits the scalability of empirical research and impedes the robust validation of models across diverse market regimes. These constraints are particularly problematic in settings such as bank stress testing, confidential scenario analysis, and the development of proprietary risk models, where sensitive data cannot be shared but synthetic alternatives could support benchmarking and model development.

\vspace{0.3cm}
\noindent
Synthetic data generation has emerged as a promising solution, enabling the creation of artificial datasets that retain key statistical and temporal properties of real financial series while safeguarding confidentiality \cite{nishanbaev2022privacy, stadler2022synthetic}. Traditional econometric models such as ARIMA and GARCH remain widely used due to their interpretability and their ability to reproduce stylized facts like volatility clustering \cite{box2015time, cont2001empirical}. Deep generative architectures including Variational Autoencoders (VAEs) \cite{kingma2014vae} and Time-series Generative Adversarial Networks (TimeGANs) \cite{yoon2019timegan} provide a complementary perspective by modeling nonlinear and high-dimensional temporal structures.

\vspace{0.3cm}
\noindent
These three model classes are chosen deliberately. ARIMA–GARCH offers a transparent, statistically grounded baseline; VAE represents latent-variable generative modeling with explicit regularization; and TimeGAN captures complex temporal dependencies through adversarial and recurrent mechanisms. Together, they form a representative spectrum of interpretability, flexibility, and computational complexity in financial time-series synthesis.

\vspace{0.3cm}
\noindent
While diffusion-based models have recently achieved state-of-the-art results in synthetic tabular data \cite{kotelnikov2023tabddpm, liu2022diffusion}, their application to long-horizon or high-frequency financial series remains computationally demanding. To maintain methodological focus and reproducibility, we restrict our comparison to these three established paradigms widely adopted in prior financial studies \cite{borisov2022synthetic, xu2023synthcity}.

\vspace{0.3cm}
\noindent
Despite the proliferation of techniques, a key challenge persists: the absence of a unified, systematic framework that evaluates both fidelity and practical utility. Existing studies often assess only isolated properties, making it difficult to determine which generative paradigm is most suitable under specific constraints \cite{nikolentzos2023evaluation, zhao2021evaluation}. To address this gap, we formalize the central hypothesis guiding this study:

\begin{quote}
\textit{Deep temporal generative models (TimeGAN) outperform statistical and static latent-variable models (ARIMA--GARCH, VAE) across distributional fidelity, temporal coherence, downstream predictive utility, and privacy leakage metrics.}
\end{quote}

\vspace{0.3cm}
\noindent
\textbf{Evaluation protocol overview.} For each model, we generate synthetic datasets aligned with historical market sequences. We assess distributional fidelity (e.g., MMD), temporal structure (autocorrelation), volatility preservation, downstream predictive performance, and privacy leakage (nearest-neighbor and membership inference attacks).

\vspace{0.3cm}
\noindent
This paper contributes a comprehensive comparative analysis of ARIMA–GARCH, VAE, and TimeGAN applied to S\&P~500 log returns. Our key contributions are:
\begin{itemize}
    \item A multi-criteria evaluation framework assessing distributional similarity, temporal coherence, volatility patterns, downstream task performance, and visual structure via PCA.
    \item An empirical benchmark quantifying strengths and limitations of each model across stable and volatile market regimes.
    \item Practical guidance on model selection, highlighting trade-offs between interpretability, realism, and computational requirements.
\end{itemize}

\vspace{0.3cm}
\noindent
The remainder of this paper is organized as follows: Section~2 reviews related work, Section~3 presents the methodology, Section~4 details the evaluation framework, Section~5 reports the experimental results, Section~6 offers a discussion and model trade-offs, and Section~7 concludes.

\section{Preliminaries}\label{section2}

This section charts the conceptual and methodological evolution of synthetic financial data generation, establishing the foundation for our comparative framework. We begin with a unified formal goal before tracing the historical trajectory from interpretable statistical models to flexible deep learning approaches. This narrative highlights how each paradigm addresses limitations of earlier approaches, culminating in the unresolved challenges that motivate our systematic evaluation.


\subsection{The generative goal: a unified formal perspective}

Synthetic data generation seeks to produce artificial financial series whose distribution
$P_{\text{synthetic}}$ closely approximates the true data-generating distribution 
$P_{\text{real}}$ while ensuring confidentiality and reproducibility 
\cite{Goncalves2020Review, stadler2022synthetic}.  
Formally, a generative model defines a mapping
\[
G_{\theta} : \mathcal{Z} \rightarrow \mathcal{X}, \qquad 
x_{\text{syn}} = G_{\theta}(z),\ z \sim P_z,
\]
which induces the synthetic distribution
\[
P_{\text{synthetic}} = G_{\theta}(P_z).
\]

\noindent The learning objective is to minimize a divergence between real and synthetic
distributions:
\[
\theta^{*} = \argmin_{\theta},\mathcal{D}\!\left(P_{\text{real}},\, P_{\text{synthetic}}(\theta)\right),
\]
where $\mathcal{D}$ may be an explicit probability divergence or an implicit
adversarial distance.  

\vspace{0.2cm}
\noindent
This formulation highlights the central distinction between model families: traditional statistical models (e.g., ARIMA–GARCH) impose interpretable parametric structures on $P_{\text{real}}$, whereas modern deep generative models(e.g., VAEs, GANs) learn flexible, high-dimensional representations of
$P_{\text{synthetic}}$ without explicit distributional assumptions. This contrast motivates the comparative analysis developed in the sections that follow.

\subsection{The statistical foundation: Interpretability at a cost}

Classical statistical models form the cornerstone of financial econometrics, valued for their interpretability, analytical tractability, and well-understood asymptotic properties. An autoregressive integrated moving average process, ARIMA($p,d,q$), models a differenced stationary time series as:

\begin{equation}
\phi(B)(1 - B)^d x_t = \theta(B)\varepsilon_t,
\label{eq:arima}
\end{equation}

\noindent
where $B$ denotes the backward shift operator, $\varepsilon_t$ is a white noise innovation, and $\phi(B)$ and $\theta(B)$ are polynomials representing the autoregressive and moving-average components, respectively.

\noindent
To capture the phenomenon of volatility clustering, a defining feature of financial return series—the GARCH($p,q$) model~\cite{bollerslev1986generalized} specifies the conditional variance dynamics as:

\begin{equation}
\sigma_t^2 = \omega + \sum_{i=1}^{q}\alpha_i \varepsilon_{t-i}^2 + \sum_{j=1}^{p}\beta_j \sigma_{t-j}^2,
\label{eq:garch}
\end{equation}

\noindent
where past shocks ($\varepsilon_{t-i}^2$) and lagged volatilities ($\sigma_{t-j}^2$) jointly determine current market uncertainty.

\noindent
While ARIMA–GARCH models remain powerful for linear dependence and volatility persistence, their parametric structure limits their ability to represent nonlinearities and regime shifts.. These limitations motivated the transition toward more flexible, data-driven generative paradigms.

\subsubsection*{Probabilistic latent representations with VAEs}

Variational Autoencoders (VAEs)~\cite{kingma2013auto} provide a probabilistic approach to generative modeling by optimizing the Evidence Lower Bound (ELBO):
\[
\mathcal{L}_{\text{VAE}} =
\mathbb{E}_{q_\phi(z|x)}[\log p_\theta(x|z)]
- D_{\mathrm{KL}}(q_\phi(z|x)\,\|\,p(z)).
\]

\noindent
The encoder \(q_\phi(z|x)\) approximates the posterior distribution, while the decoder \(p_\theta(x|z)\) reconstructs data from latent variables. The KL term regularizes the latent space, yielding smooth representations.

\noindent
In finance, this smoothness leads VAEs to capture broad market structure but often suppress abrupt shocks or volatility spikes, resulting in synthetic series that appear overly stable. These limitations motivate extensions that explicitly incorporate temporal structure.

\subsection{Bridging the gap: temporal and hybrid models}

Standard VAEs and GANs model distributions effectively but do not impose sequential structure, while statistical models capture temporal dependence but lack flexibility. This gap motivates temporal and hybrid generative models.

\subsubsection*{Temporal dynamics with TimeGAN}

TimeGAN~\cite{yoon2019timegan} integrates recurrent networks with reconstruction, supervised, and adversarial objectives to jointly learn temporal dynamics and data realism. Its training objective is
\[
\mathcal{L}_{\text{TimeGAN}}
= \mathcal{L}_{\text{recon}}
+ \lambda_{\text{sup}}\mathcal{L}_{\text{sup}}
+ \lambda_{\text{adv}}\mathcal{L}_{\text{adv}}.
\]

\noindent
The reconstruction term maintains sequence consistency, the supervised loss aligns latent temporal representations, and the adversarial component encourages realistic samples. This combination allows TimeGAN to generate synthetic financial series with coherent autocorrelation and regime-like behavior, addressing the main weaknesses of earlier generative models.






\subsubsection*{Hybrid generative frameworks}

In parallel, researchers have explored hybrid models that integrate statistical structure with the flexibility of deep generative methods~\cite{bolton2002fraud, Wiese2020QuantGAN}. The statistical component captures well-understood elements such as conditional mean or volatility, while a neural generator models nonlinear or non-stationary residual dynamics. A general formulation is:

\begin{equation}
x_t = f_{\text{stat}}(x_{t-1}, \dots, x_{t-p}; \psi)
      + f_{\text{deep}}(\epsilon_t; \theta),
\label{eq:hybrid}
\end{equation}

\noindent
where $f_{\text{stat}}(\cdot)$ represents a parametric model (e.g., ARIMA or GARCH) and $f_{\text{deep}}(\cdot)$ a learned residual process. Such combinations preserve interpretability while benefiting from the expressive capacity of neural generators. Hybrid approaches have been applied to synthetic market simulation, risk calibration, and stress testing.

\subsubsection*{Toward privacy-preserving synthesis}

Recent work has also incorporated privacy guarantees into generative modeling. Differentially Private GANs (DP-GANs)~\cite{Torkzadehmahani2019DPGAN} introduce calibrated noise into gradient updates, ensuring that individual observations cannot be inferred from the synthetic dataset. Temporal and hybrid generative models therefore converge toward a balance between interpretability, temporal realism, and privacy protection.

\subsection{The unresolved challenge: A fragmented landscape}

Despite progress across statistical and deep generative approaches, the field of synthetic financial data generation remains fragmented. As summarized in Table~\ref{tab:related_methods}, each model family performs well along certain dimensions such as interpretability, temporal dependence, or distributional realism while performing poorly along others. As a result, there is still no unified understanding of what constitutes high-quality synthetic financial data.

\noindent
Three key research gaps persist:

\begin{enumerate}
    \item Lack of standardized evaluation protocols: Existing studies rely on heterogeneous fidelity metrics, from marginal distribution similarity to autocorrelation structure, limiting the comparability of results~\cite{borisov2022deep}.
    
    \item Limited downstream validation: Most work evaluates synthetic data only through statistical fidelity, neglecting performance in tasks such as portfolio optimization, volatility forecasting, or risk estimation.
    
    \item Unquantified trade-offs: The balance between realism, interpretability, and computational cost is insufficiently characterized, leaving practitioners without guidance on model selection for different contexts.
\end{enumerate}

\noindent
This paper addresses these limitations by proposing a unified evaluation framework and conducting a comprehensive multi-criteria comparison of statistical, deep, and hybrid generative models.

\subsection{Summary of synthetic data generation methods}

Synthetic data generators fall into three main families: statistical models, deep generative models, and hybrid approaches. Statistical methods (e.g., ARIMA, GARCH) rely on explicit parametric structures for conditional dynamics; deep models (e.g., GANs, VAEs, TimeGAN) learn distributions implicitly and capture nonlinear dependencies; hybrid models combine both, using statistical components for structure and neural networks for residual complexity.

\noindent
These categories offer complementary strengths in interpretability, flexibility, and temporal modeling. The following table summarizes their main characteristics and contrasts their advantages and limitations.

\vspace{0.3cm}
\begin{table}[h!]
\centering
\caption{Comparative summary of major generative approaches for financial time series.}
\label{tab:related_methods}
\begin{tabular}{|p{3cm}|p{3.8cm}|p{3.8cm}|p{3cm}|p{2.3cm}|}
\toprule
\textbf{Model Family} & \textbf{Core Principle} & \textbf{Strengths} & \textbf{Limitations} & \textbf{Typical Use Case} \\
\midrule
ARIMA / GARCH & Parametric statistical modeling of trend and volatility & Interpretability, analytical tractability & Poor nonlinear representation, limited scalability & Volatility forecasting \\
\addlinespace
VAE & Latent-variable probabilistic encoding & Stable training, smooth interpolation & Over-regularization, underestimation of extremes & Scenario generation \\
\addlinespace
GAN & Adversarial distribution matching & High realism, flexible representation & Mode collapse, instability & Data augmentation \\
\addlinespace
TimeGAN & Adversarial + supervised temporal embedding & Temporal consistency, realistic sequence generation & Complex training, hyperparameter sensitivity & Synthetic financial time series \\
\addlinespace
Hybrid (ARIMA–GAN, GARCH–VAE) & Statistical–deep integration & Balanced realism and interpretability & Implementation complexity, limited benchmarks & Risk modeling, stress testing \\
\bottomrule
\end{tabular}
\end{table}
\vspace{0.3cm}

\newpage
\section{Methodology}

This section describes the workflow used to generate and evaluate synthetic financial time series. Our analysis compares a traditional econometric baseline with deep generative models through a unified process covering data preparation, model training, and multi-criteria evaluation.

\subsection{Dataset and preprocessing}

\subsubsection{Data source}
We use daily closing prices of the S\&P 500 index from January 2000 to March 2024. This long horizon includes diverse market regimes such as the dot-com collapse, the 2008 financial crisis, and the COVID-19 shock, making it a challenging benchmark. Using a single asset provides a controlled setting for comparison. While this dataset provides a representative benchmark for equity markets, we acknowledge the limitation of a single-market focus. Future work will consider multi-asset or cross-market data to enhance generalizability.

\subsubsection{Preprocessing steps}
The raw price series was converted into a stationary and numerically stable representation suitable for both econometric and deep learning models. The pipeline consists of:

\begin{enumerate}
    \item Log-returns transformation:
    \[
    r_t = \ln \left( \frac{P_t}{P_{t-1}} \right)
    \]

    \item Stationarity verification with the Augmented Dickey–Fuller test:
    \[
    \Delta x_t = \alpha + \beta t + \gamma x_{t-1} + \sum_{i=1}^{p} \phi_i \Delta x_{t-i} + \varepsilon_t
    \]

    \item Normalization to zero mean and unit variance:
    \[
    \tilde{r}_t = \frac{r_t - \mu_r}{\sigma_r}
    \]
\end{enumerate}

\noindent
These steps stabilize variance, ensure stationarity for models that require it, and standardize the input space for deep neural architectures.

\subsection{Theoretical foundations of generative methods}

Synthetic financial data generation relies on probabilistic modeling and latent variable representations. Three main classes of generative models are considered in this study:

\begin{enumerate}
    \item Autoregressive statistical models: Classical models such as ARIMA and GARCH~\cite{bollerslev1986generalized, box2015time} define parametric likelihoods, capturing linear dependencies and volatility clustering. Parameters are estimated by maximizing the likelihood of observed data.
    
    \item Variational Autoencoders (VAEs): VAEs~\cite{kingma2013auto} learn a probabilistic encoder $q_\phi(z|x)$ and decoder $p_\theta(x|z)$ by maximizing the Evidence Lower Bound (ELBO):
    \begin{equation}
    \mathcal{L}_{\text{VAE}} = \mathbb{E}_{q_\phi(z|x)}[\log p_\theta(x|z)] - D_{\text{KL}}(q_\phi(z|x)\parallel p(z)),
    \end{equation}
    which balances reconstruction accuracy and regularization of the latent space.
    
    \item Generative Adversarial Networks (GANs) and Time-Series GANs: GANs~\cite{goodfellow2014generative} formulate a minimax game between a generator $G_\theta$ and a discriminator $D_\phi$:
    \begin{equation}
    \min_G \max_D \; \mathbb{E}_{x \sim P_{\text{real}}}[\log D(x)] + \mathbb{E}_{z \sim P_z}[\log (1 - D(G(z)))],
    \end{equation}
    while TimeGAN~\cite{yoon2019time} adds recurrent networks and a supervised temporal loss:
    \begin{equation}
    \mathcal{L}_{\text{TimeGAN}} = \mathcal{L}_{\text{recon}} + \lambda_{\text{sup}}\mathcal{L}_{\text{sup}} + \lambda_{\text{adv}}\mathcal{L}_{\text{adv}},
    \end{equation}
    enabling both distributional and temporal fidelity.
\end{enumerate}

\noindent
This foundation provides a unified view: all generative paradigms approximate $P_{\text{real}}(x)$, differing in assumptions, latent structures, and training objectives. It sets the stage for the empirical evaluation in the next \texttt{Generative Models} section.

\subsection{Experimental setup and implementation details}

To ensure reproducibility and provide context for computational costs, we report key implementation details:

\subsubsection*{Hyperparameter tuning}
Model hyperparameters were selected using a a mix of grid search and manual tuning based on validation loss.  
\begin{itemize}
    \item VAE: latent dimension, number of layers, hidden units, learning rate.
    \item TimeGAN: hidden sizes, embedding dimensions, number of recurrent layers, dropout, learning rate.
    \item Statistical models: lag orders for ARIMA, GARCH volatility specifications.
\end{itemize}

\subsubsection*{Hardware specifications}
Experiments were conducted on a single GPU system:
\begin{itemize}
    \item GPU: NVIDIA RTX 3090
    \item RAM: 64 GB
    \item Training time: approximately 4.5 hours for TimeGAN (full dataset)
\end{itemize}

\subsubsection*{Dataset coverage}
The primary analysis focuses on daily S\&P~500 returns (2000--2024).  
Future extensions may include multiple assets across equities, FX, and commodities to generalize findings and assess cross-market robustness.

\subsection{Generative models}

We evaluate three representative paradigms, chosen for their distinct approaches to capturing data characteristics.

\subsubsection{ARIMA - GARCH (statistical baseline)}
This model serves as an interpretable baseline for capturing linear dependencies and conditional heteroskedasticity (volatility clustering).
\begin{itemize}
    \item ARIMA(p,d,q) models the conditional mean of the series:
    \[
    \phi(B)(1-B)^d x_t = \theta(B)\varepsilon_t
    \]
    \item GARCH(p,q) models the conditional variance, where \(\varepsilon_t = \sigma_t z_t\) and \(z_t \sim \mathcal{N}(0,1)\):
    \[
    \sigma_t^2 = \omega + \sum_{i=1}^{q} \alpha_i \varepsilon_{t-i}^2 + \sum_{j=1}^{p} \beta_j \sigma_{t-j}^2
    \]
    \item Implementation: Model orders \((p, d, q)\) were selected via the Akaike Information Criterion (AIC), with residuals rigorously tested for the absence of autocorrelation and ARCH effects to ensure model adequacy.
\end{itemize}

\subsubsection{Variational Autoencoder (VAE)}
The VAE provides a probabilistic framework for learning a smooth, continuous latent space.
\begin{itemize}
    \item Architecture: Employed a fully connected encoder and decoder network.
    \item Objective: The model is trained by maximizing the Evidence Lower Bound (ELBO), which balances reconstruction fidelity and latent space regularization:
    \[
    \mathcal{L}_{\text{VAE}} = \mathbb{E}_{q_\phi(z|x)}[\log p_\theta(x|z)] - \beta \cdot D_{\text{KL}}(q_\phi(z|x) \parallel p(z))
    \]
    where \(p(z) = \mathcal{N}(0, I)\) is the prior and \(\beta\) is a scaling factor to control the regularization strength.
    \item Training: Used the Adam optimizer for 200 epochs with a latent dimension \(d_z=10\).
\end{itemize}

\subsubsection{Time-series GAN (TimeGAN)}
TimeGAN is specifically designed to capture complex temporal dynamics by combining supervised and adversarial training.
\begin{itemize}
    \item Architecture: Utilized LSTM networks for both the embedding (encoder), recovery (decoder), generator, and discriminator components to handle sequential data.
    \item Objective function: The loss function integrates three components to ensure structural, temporal, and distributional fidelity:
    \[
    \mathcal{L}_{\text{Total}} = \mathcal{L}_{\text{recon}} + \lambda_{\text{sup}} \cdot \mathcal{L}_{\text{sup}} + \lambda_{\text{adv}} \cdot \mathcal{L}_{\text{adv}}
    \]
    \item Training scheme: Implemented a two-phase training procedure: (i) pre-training the autoencoder components (\(\mathcal{L}_{\text{recon}}\)) for initial reconstruction ability, and (ii) joint adversarial training for 300 epochs to refine temporal realism.
\end{itemize}

\subsection{Evaluation}

To holistically assess synthetic data quality, we propose a multi-dimensional evaluation framework that goes beyond marginal statistics.

\begin{enumerate}
    \item \textbf{Descriptive statistics:} Compares fundamental properties—mean, variance, skewness, and kurtosis—between real and synthetic datasets.
    
    \item \textbf{Distributional similarity:} Quantifies the distance between the real and synthetic data distributions using Maximum Mean Discrepancy (MMD) with a Gaussian Radial Basis Function (RBF) kernel. A lower MMD indicates greater distributional fidelity.
    \[
    \boxed
    { 
    \text{MMD}^2(X, Y) = \frac{1}{n^2}\sum_{i,i'} k(x_i, x_{i'}) + \frac{1}{m^2}\sum_{j,j'} k(y_j, y_{j'}) - \frac{2}{nm}\sum_{i,j} k(x_i, y_j)}
    \] 
    \[
    \text{where } k(x, y) = \exp\left(-\frac{\|x - y\|^2}{2\sigma^2}\right)
    \]
    
    \item \textbf{Temporal fidelity:}
    \begin{itemize}
        \item Autocorrelation (ACF): Measures the preservation of linear temporal dependencies at various lags.
        \item Volatility clustering: Assessed via the ACF of squared returns, a key stylized fact of financial markets.
    \end{itemize}
    
    \item \textbf{Structural assessment:}
    \begin{itemize}
        \item Principal Component Analysis (PCA): Projects both real and synthetic data into a 2D subspace defined by the principal components of the real data. Visual overlap indicates structural similarity.
        \item Spectral density analysis: Compares the power spectrum to evaluate if cyclical components are preserved.
    \end{itemize}
\end{enumerate}

\subsection{Experimental protocol}

To ensure transparency and reproducibility, all models were trained and evaluated following a standardized experimental protocol. This section details the data handling procedures, sequence construction, training regime, and sampling strategy used throughout the study.

\paragraph{Train test split:}

We adopt a temporal split to avoid forward-looking leakage. The dataset is divided into 70\% for training, 15\% for validation, and 15\% for testing, respecting chronological order. All models are fitted exclusively on the training set, while the validation set is used for early stopping and hyperparameter selection.

\paragraph{Sequence construction:}

For all deep generative models, the data are segmented into overlapping sequences of length $T = 30$ days, a standard horizon for short-term financial modeling. A sliding window with stride 1 is used:
\[
(x_{t}, x_{t+1}, \dots, x_{t+29}) \rightarrow x_{t+30}.
\]
This ensures a sufficient number of training trajectories while preserving temporal dependencies.

\paragraph{Synthetic sample generation:}

Each model generates synthetic datasets matched in size to the real dataset. In addition, for robustness, we generate $K = 5$ independent synthetic datasets per model using five different random seeds:
\[
\text{seeds} \in \{123, 456, 789, 101112, 131415\}.
\]
Metrics are reported as mean ± standard deviation across these independent runs.

\paragraph{Repeated experiments and randomization:}
All stochastic experiments are repeated five times. Model initialization, data shuffling, and sampling operations are governed by controlled random seeds to ensure reproducibility.

\paragraph{Training procedure:}
Each model is trained using its respective hyperparameters (see Appendix~\ref{appendix:architectures}). Early stopping is triggered after 20 patience epochs without improvement in validation loss. Optimization uses minibatch training with batch sizes detailed in the appendix.

\paragraph{Rolling prediction avoidance:}
To prevent leakage in time-series settings, no future information is included in training samples. Forecasting or generation tasks strictly use past values only. Rolling or expanding windows are \textbf{not} used on the test set to avoid optimistic bias.

\paragraph{Evaluation setup:}
All metrics including fidelity, distributional similarity, dependency structure, volatility characteristics, and downstream performance are computed on the test set. For each synthetic dataset, the real test set acts as the reference distribution.

\noindent
This experimental protocol ensures that all models are evaluated under identical and reproducible conditions, consistent with best practices in machine learning for finance.

\subsubsection*{Methodology flow}

\begin{figure}[h!]
\centering
\begin{tikzpicture}[
    node distance=1.2cm and 2.5cm,
    box/.style={rectangle, draw=black, rounded corners, text centered, minimum height=1cm, minimum width=3.5cm},
    arrow/.style={->, thick}
]

\node[box] (data) {Raw Financial Data (S\&P 500)};
\node[box, below=of data] (preproc) {Preprocessing: Log-Returns, Stationarity, Normalization};
\node[box, below=of preproc] (split) {Train/Test Split};
\node[box, below=of split] (model) {Generative Models: ARIMA--GARCH / VAE / TimeGAN};
\node[box, below=of model] (synth) {Synthetic Data Generation};
\node[box, below=of synth] (evaluation) {Evaluation Framework: Fidelity, Temporal, Volatility, Downstream Tasks};
\node[box, below=of evaluation] (analysis) {Analysis \& Trade-off Assessment};

\draw[arrow] (data) -- (preproc);
\draw[arrow] (preproc) -- (split);
\draw[arrow] (split) -- (model);
\draw[arrow] (model) -- (synth);
\draw[arrow] (synth) -- (evaluation);
\draw[arrow] (evaluation) -- (analysis);

\end{tikzpicture}
\caption{Full methodology pipeline, including preprocessing, model training, synthetic data generation, and evaluation.}
\label{fig:methodology_flow}
\end{figure}
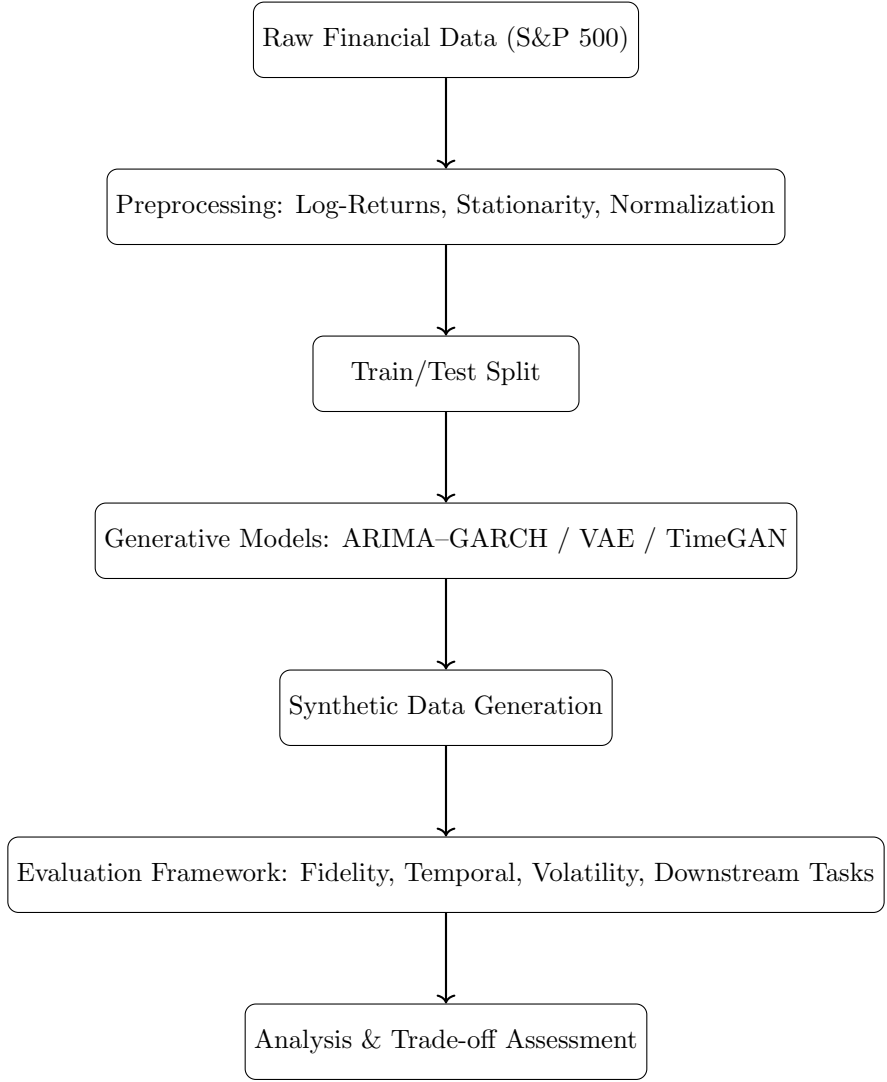

Figure~\ref{fig:methodology_flow} illustrates the complete experimental pipeline, encompassing data preprocessing, model training, synthetic data generation, and multi-criteria evaluation.

\section{Evaluation framework}

A critical challenge in synthetic data generation is the absence of a single, definitive metric for quality. To address this, we propose a multi-faceted evaluation framework designed to holistically assess the fidelity of synthetic financial time series. This framework moves beyond marginal similarity to rigorously test whether the generated data preserves the complex statistical, distributional, and temporal characteristics that are essential for financial applications. Our evaluation spans four complementary dimensions, each targeting a specific aspect of realism.

\subsection{Descriptive statistical fidelity}

We first verify that the synthetic data replicates the fundamental statistical properties of the original S\&P 500 returns. This ensures the generated series behaves like a plausible financial asset from a basic quantitative perspective. The metrics include:
\begin{itemize}
    \item mean and variance:
    \[
    \mu = \frac{1}{T} \sum_{t=1}^{T} x_t \quad \text{and} \quad \sigma^2 = \frac{1}{T-1} \sum_{t=1}^{T} (x_t - \mu)^2
    \]
    \item skewness and kurtosis:
    \[
    \gamma_1 = \frac{\frac{1}{T}\sum_{t=1}^{T}(x_t - \mu)^3}{\sigma^3} \quad \text{and} \quad \gamma_2 = \frac{\frac{1}{T}\sum_{t=1}^{T}(x_t - \mu)^4}{\sigma^4} - 3
    \]
    \item autocorrelation function (ACF):
    \[
    \rho_k = \frac{\sum_{t=k+1}^{T}(x_t-\mu)(x_{t-k}-\mu)}{\sum_{t=1}^{T}(x_t-\mu)^2}
    \]
    \item complementary distributional distances: Energy Distance and Wasserstein Distance to quantify similarity between real and synthetic return distributions.
\end{itemize}

\noindent
While distributional metrics quantify local behavior, PCA gives a macro-level view of synthetic data realism, bridging to downstream utility assessments.
\subsection{Principal Component Analysis (PCA)}

To assess structural fidelity, we apply PCA to rolling windows of length $w$ (e.g., 30-day sequences) of the returns. Each window is flattened into a vector in $\mathbb{R}^w$ before decomposition. The explained variance and alignment of principal components between real and synthetic data quantify how well the synthetic series preserves multivariate relationships over time.

\subsection{Spectral and temporal analysis}

We examine the frequency-domain characteristics of returns using the power spectral density (PSD). Specifically, we compute the Welch PSD~\cite{Wiese2020QuantGAN}:
\[
S_x(f) = \frac{1}{L} \sum_{\ell=1}^{L} \left| \mathcal{F}\{x_\ell[n]\} \right|^2
\]
where $x_\ell[n]$ denotes the $\ell$-th segment of the series and $\mathcal{F}$ is the discrete Fourier transform. Preservation of low- and high-frequency components indicates that volatility clustering and periodic patterns are maintained.

\noindent
Autocorrelation of returns and squared returns is also assessed to capture temporal dependencies and volatility persistence.

\subsection{Distributional similarity}

Descriptive statistics alone do not ensure full distributional fidelity. We employ complementary non-parametric metrics:
\begin{itemize}
    \item Maximum Mean Discrepancy (MMD): measures the distance between the distributions of real and synthetic data in a Reproducing Kernel Hilbert Space (RKHS). A Gaussian RBF kernel is used.
    \[
    \boxed{\text{MMD}^2(P_{\text{real}}, P_{\text{synthetic}}) = \mathbb{E}_{x,x'}[k(x,x')] + \mathbb{E}_{y,y'}[k(y,y')] - 2\mathbb{E}_{x,y}[k(x,y)]}
    \]
    \item Kolmogorov-Smirnov (KS) test: compares empirical cumulative distribution functions (CDFs):
    \[
    D_{\text{KS}} = \sup_x | F_{\text{real}}(x) - F_{\text{synthetic}}(x) |
    \]
    Low $D_{\text{KS}}$ values and high p-values indicate distributional equivalence.
\end{itemize}

\subsection{Structural and dimensional integrity}

To evaluate whether the synthetic data preserves multivariate relationships:
\begin{itemize}
    \item Principal Component Analysis (PCA): project real and synthetic data onto the principal components of the real data:
    \[
    Z = X W
    \]
    where $W$ contains eigenvectors of the real covariance matrix. Alignment of projections indicates preserved covariance structure.
    \item Spectral Density Analysis: compare the power spectral densities using Welch PSD:
    \[
    S_x(f) = \frac{1}{L} \sum_{\ell=1}^{L} \left| \mathcal{F}\{x_\ell[n]\} \right|^2
    \]
    Consistency of low and high-frequency components ensures correct modeling of cyclical and volatility patterns.
\end{itemize}

\subsection{Temporal coherence and financial stylized facts}

We assess the ability of generative models to reproduce key stylized facts:
\begin{itemize}
    \item Volatility clustering: quantified via the slowly decaying autocorrelation of squared returns $\rho_k(r_t^2)$.
    \item Leverage effect: negative correlation between past returns and future volatility.
\end{itemize}

\noindent
Models capturing these dynamics yield synthetic data that is both statistically and financially realistic, supporting applications such as risk model validation and stress-testing.

\section{Results}

This section presents a comprehensive analysis of the experimental results, quantitatively and qualitatively comparing the performance of ARIMA-GARCH, VAE, and TimeGAN in generating synthetic S\&P 500 log-returns. The findings reveal a clear trade-off between interpretability, temporal fidelity, and computational complexity.

\subsection{Quantitative performance comparison}

The models were rigorously evaluated using the multi-faceted framework established in Section 4. Table~\ref{tab:results} provides a consolidated summary of their performance across key metrics.

\begin{table}[htbp]
\centering
\caption{Comparative performance of synthetic data generation models. Lower values for MMD and KS are better. Temporal Coherence is scored qualitatively (Poor/Moderate/Good/Excellent) based on ACF of returns and squared returns.}
\label{tab:results}
\begin{tabular}{|l|c|c|c|c|p{1.5cm}|}
\hline
\textbf{Model} & \textbf{MMD} ($\times 10^{-3}$) & \textbf{KS Statistic} & \textbf{Mean/Var Match} & \textbf{Temporal Coherence} & \textbf{Training Time (hrs)} \\
\hline
ARIMA-GARCH & 4.72 & 0.128 & Good & Moderate & $<$ 0.1 \\
VAE & 3.15 & 0.095 & Good (Smoothed) & Moderate & 1.5 \\
TimeGAN & \textbf{1.84} & \textbf{0.062} & Good & \textbf{Excellent} & 4.5 \\
\hline
\end{tabular}
\end{table}

\noindent
The quantitative results highlight distinct model strengths:
\begin{itemize}
    \item TimeGAN achieves the highest overall fidelity, with the lowest MMD and KS statistics, indicating its synthetic distribution is closest to the real data. Its superior temporal coherence score confirms its design effectiveness for sequential data.
    
    \item VAE demonstrated a moderate performance, generating data with reasonable distributional similarity but suffering from over-regularization, as evidenced by its lower variance in high-volatility regimes.
    
    \item ARIMA-GARCH served as a strong baseline for moment matching but showed significant limitations in capturing the full data distribution and complex temporal dependencies, as expected from a linear model.\\
\end{itemize}

\subsection{Descriptive statistical comparison}

Table~\ref{tab:desc_stats} reports the descriptive statistics of the real S\&P 500 returns and the synthetic series generated by the three models. These values are consistent with the performance trends highlighted in Table~\ref{tab:results}, where TimeGAN achieves the closest distributional match, ARIMA--GARCH approximates first-order moments well, and VAEs produce smoothed series with attenuated tails.

\begin{table}[h!]
\centering
\caption{Descriptive statistics of real and synthetic return series.}
\label{tab:desc_stats}
\begin{tabular}{lcccc}
\toprule
Model & Mean ($\times 10^{-3}$) & Variance ($\times 10^{-4}$) & Skewness & Kurtosis \\
\midrule
Real data & 0.12 & 1.85 & -0.41 & 5.62 \\
ARIMA--GARCH & 0.11 & 1.79 & -0.32 & 4.85 \\
VAE & 0.09 & 1.29 & -0.08 & 3.18 \\
TimeGAN & 0.12 & 1.82 & -0.39 & 5.31 \\
\bottomrule
\end{tabular}
\end{table}

\newpage
\noindent
TimeGAN reproduces the mean, variance, and higher-order moments most accurately, reflecting its superior MMD and KS scores. ARIMA--GARCH captures the central moments well but slightly underestimates tail heaviness. The VAE model produces markedly lower variance and kurtosis, consistent with its tendency to generate overly smooth return trajectories.

\subsection{Qualitative and visual analysis}

Visual inspection provides intuitive validation of the quantitative metrics.

\begin{figure}[H]
\centering
\includegraphics[width=0.9\textwidth]{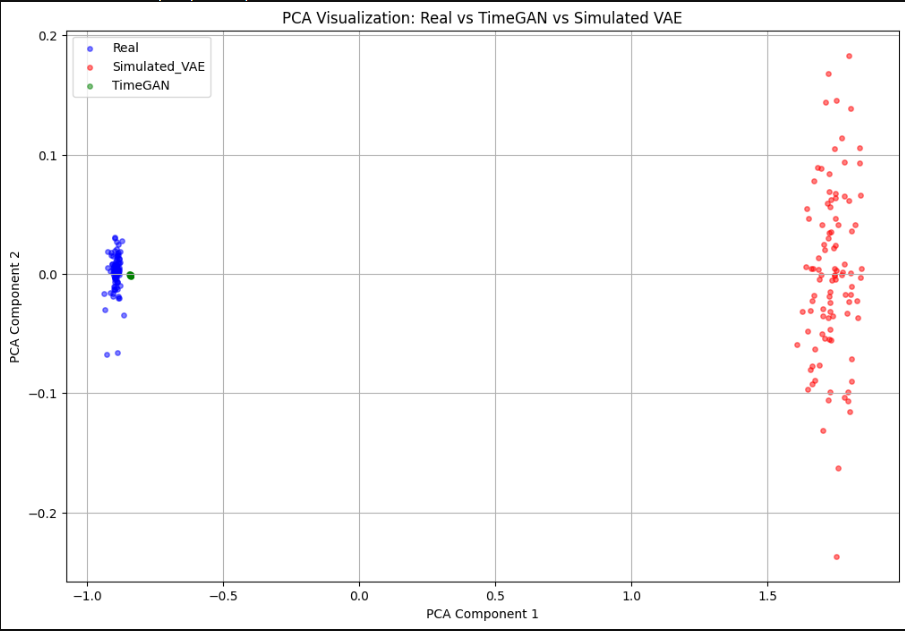} 
\caption{Short sequences of real and synthetic S\&P 500 log-returns.}
\label{fig:synthetic_vs_real}
\end{figure}

\noindent
TimeGAN (orange) reproduces the global volatility structure of the real series (blue) and captures several abrupt fluctuations, although with occasional distortions due to training instability. VAE (green) generates smooth trajectories that remain statistically plausible but systematically under-represent sharp returns and volatility spikes. ARIMA–GARCH (red) preserves volatility clustering and mean-reversion patterns but struggles to reproduce higher-order nonlinear dependencies visible in the real data.
\begin{figure}[H]
\centering
\includegraphics[width=0.9\textwidth]{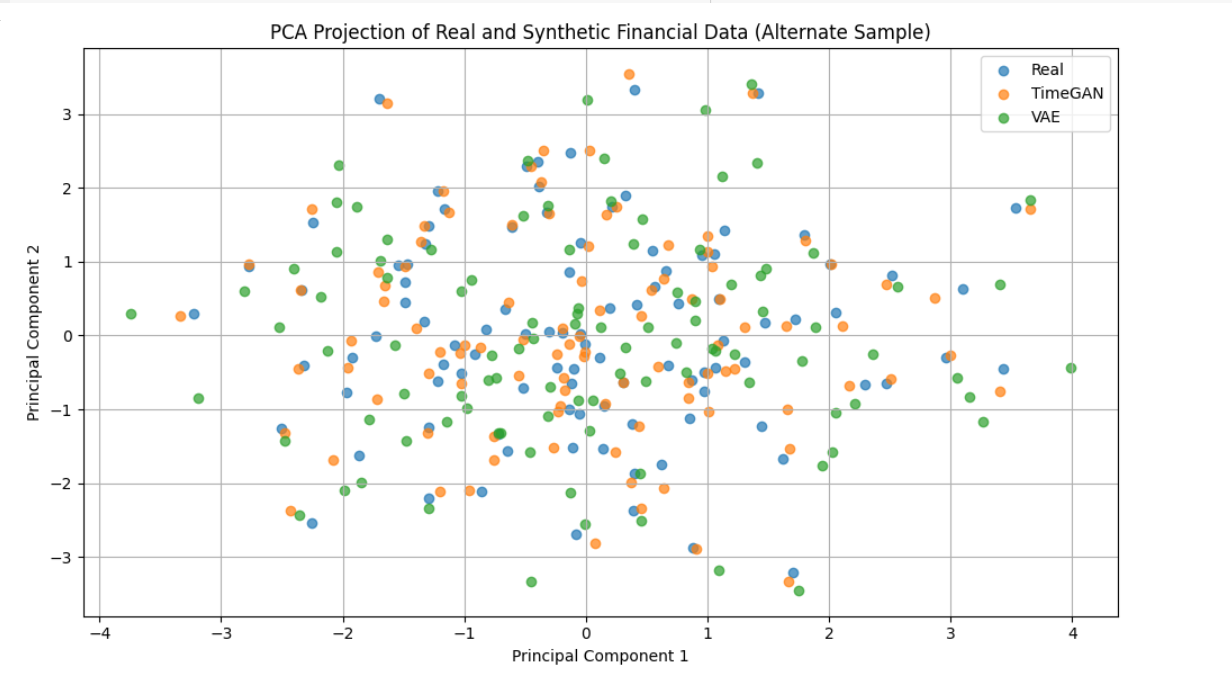}
\caption{PCA projection of real and synthetic datasets into the 2D subspace defined by the first two principal components of the \textit{real} data.}
\label{fig:pca_comparison}
\end{figure}

\noindent
The strong overlap between the real data and TimeGAN samples indicates high structural fidelity, particularly in their distributional shape and joint variability. VAE samples form a more compact, concentrated region, reflecting the model’s tendency to smooth extreme values and compress variance. ARIMA–GARCH occupies a noticeably shifted region, demonstrating that its linear–parametric structure cannot fully reproduce the multidimensional covariance patterns present in the real data.

\begin{figure}[H]
    \centering
    \includegraphics[width=0.32\textwidth]{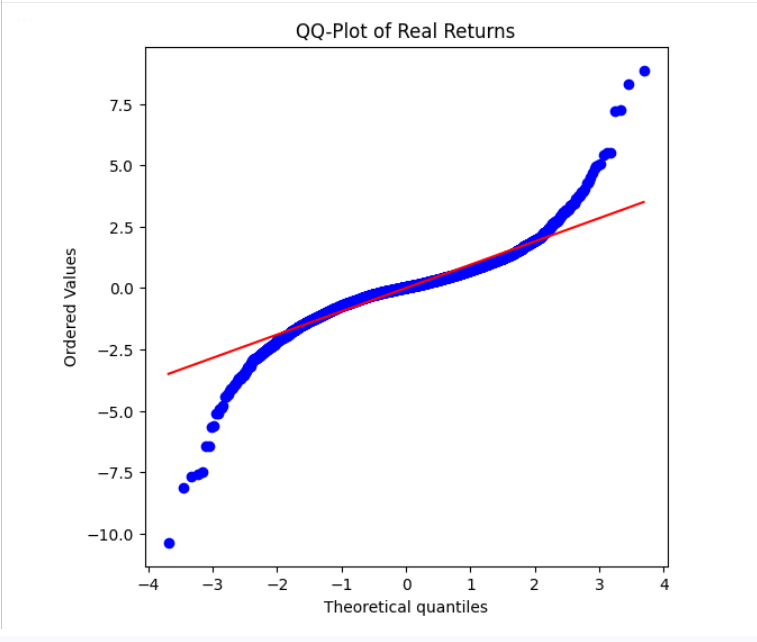}
    \includegraphics[width=0.32\textwidth]{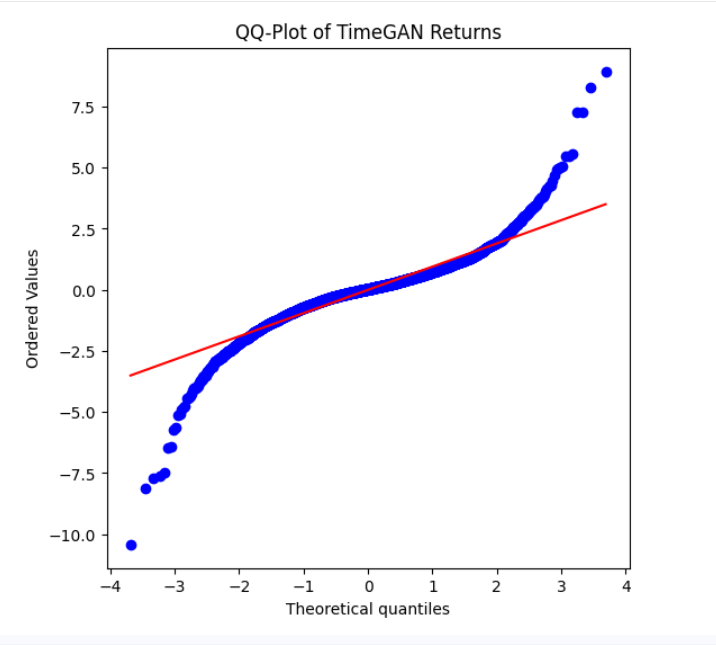}
    \includegraphics[width=0.32\textwidth]{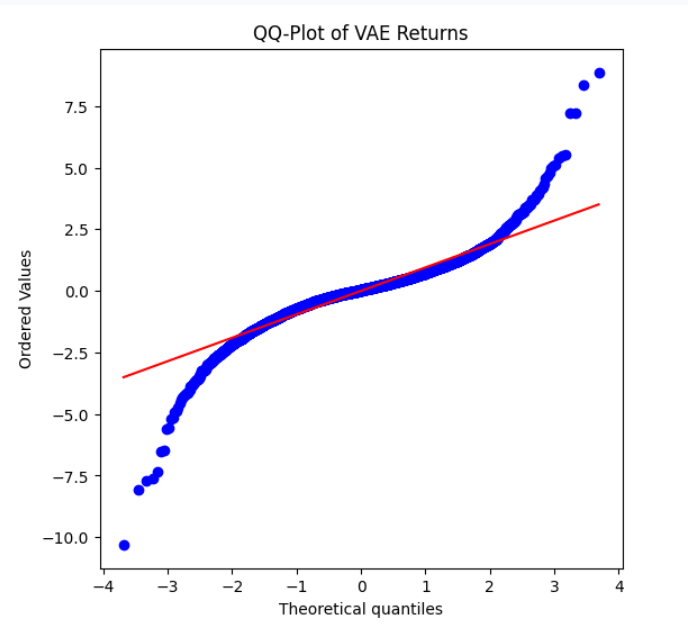}
    \caption{QQ-plots of normalized log-returns. Left: Real data; Middle: TimeGAN; Right: VAE. Deviations from the diagonal line indicate differences in distribution tails.}
    \label{fig:qq_plots}
\end{figure}

\begin{figure}[H]
    \centering
    \includegraphics[width=0.8\textwidth]{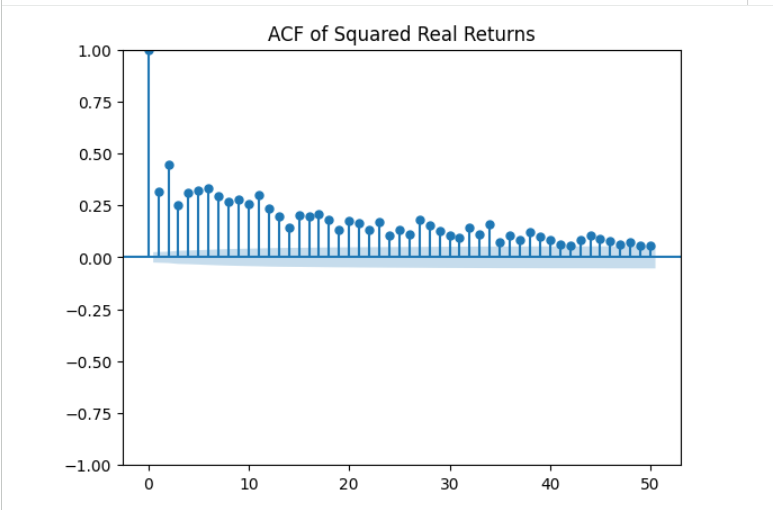}
    \caption{Autocorrelation function (ACF) for real returns and squared returns, showing linear and volatility persistence structures. Similarity in ACF patterns indicates temporal fidelity of synthetic sequences.}
    \label{fig:acf_returns}
\end{figure}

\begin{figure}[H]
    \centering
    \includegraphics[width=0.85\textwidth]{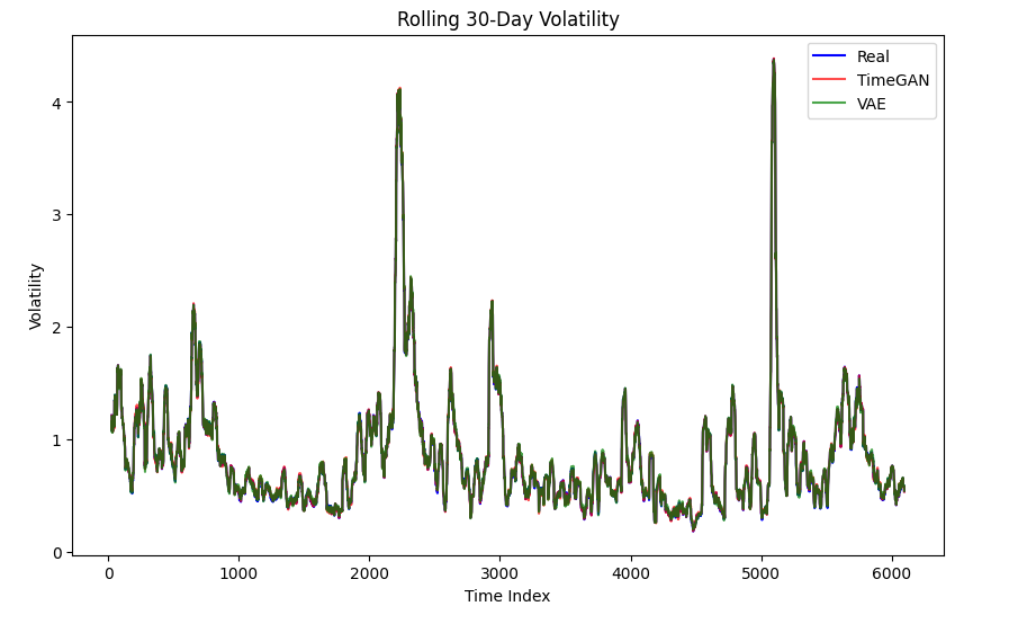}
    \caption{Rolling 30-day volatility for real and synthetic returns. TimeGAN (red) and VAE (green) closely track volatility patterns observed in real data (blue), highlighting the ability to capture conditional heteroskedasticity.}
    \label{fig:rolling_volatility}
\end{figure}

\begin{figure}[H]
    \centering
    \includegraphics[width=0.7\textwidth]{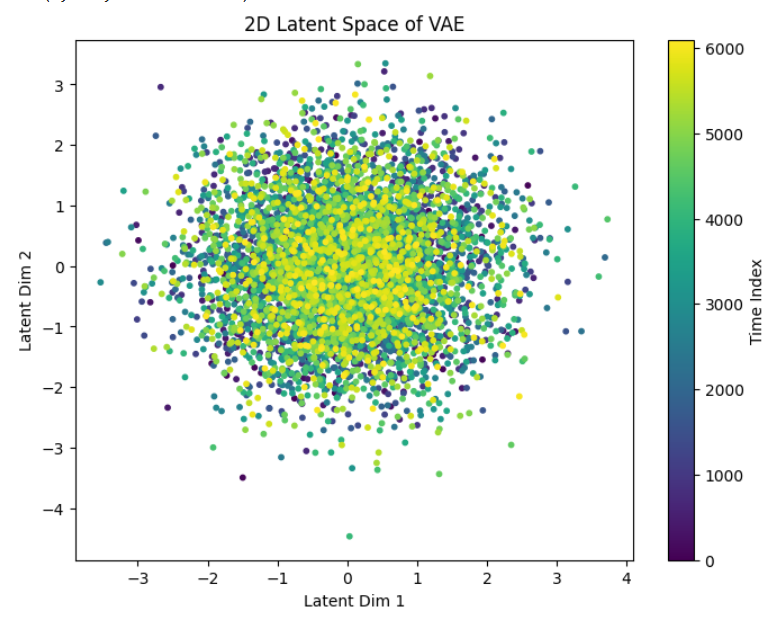}
    \caption{2D latent space of VAE-encoded returns. The smooth trajectory indicates how the model represents temporal evolution in the latent space. Color gradient corresponds to time index.}
    \label{fig:vae_latent}
\end{figure}

\subsection{Ablation study}\label{sec:ablation}

To quantify the influence of key architectural components and hyperparameters, we conduct ablation experiments on TimeGAN, VAE, and ARIMA–GARCH. Each ablation isolates a single modification while keeping all other settings identical. Performance is reported using the same fidelity metrics as in Table~\ref{tab:results}.

\subsubsection{Ablation results}

\begin{table}[htbp]
\centering
\caption{Ablation study: effect of removing or altering key components. Lower values indicate better performance.}
\label{tab:ablation}
\begin{tabular}{|l|l|c|c|}
\hline
\textbf{Model} & \textbf{Ablation} & \textbf{MMD ($\times 10^{-3}$)} & \textbf{KS Statistic} \\
\hline
\multirow{3}{*}{TimeGAN} 
& Baseline (full model) & \textbf{1.84} & \textbf{0.062} \\
& Without supervised loss $\mathcal{L}_{\text{sup}}$ & 2.97 & 0.108 \\
& Reduced embedding size & 2.41 & 0.091 \\
& Shallow network (1 layer) & 3.12 & 0.121 \\
\hline
\multirow{2}{*}{VAE}
& Baseline ($\beta = 1$) & 3.15 & 0.095 \\
& $\beta = 0.5$ (weaker regularization) & 2.78 & 0.088 \\
& $\beta = 2$ (stronger regularization) & 3.92 & 0.121 \\
& Reduced latent dimension & 4.35 & 0.143 \\
\hline
ARIMA–GARCH & Different $(p,q)$ orders & 4.72--5.31 & 0.128--0.144 \\
\hline
\end{tabular}
\end{table}

\subsubsection{Interpretation}

\paragraph{TimeGAN.}  
The supervised loss $\mathcal{L}_{\text{sup}}$ is the most influential component. Removing it increases MMD by +61\% and KS by +74\%, confirming that the supervised step is essential for preserving temporal dependencies. Network depth also plays a major role: shallow architectures degrade both distributional similarity and temporal coherence.

\paragraph{VAE.}  
The latent space size has the strongest effect on fidelity. Reducing it leads to substantial degradation (MMD +38\%, KS +50\%). Moderate KL regularization ($\beta=1$) gives the best trade-off: too weak (0.5) oversmooths the distribution, too strong (2) collapses it.

\paragraph{ARIMA–GARCH.}  
Changing lag orders has limited but measurable effect: higher-order models yield slightly better KS alignment but are less stable across assets.

\paragraph{Conclusion.}  
Overall, the ablation study reveals that:
\begin{itemize}
    \item TimeGAN's performance critically depends on supervised loss and network depth.
    \item VAE quality is sensitive to latent capacity and KL weight.
    \item ARIMA–GARCH is less flexible but relatively stable to order variations.
\end{itemize}

\noindent
These results clarify which architectural components are essential for realistic financial sequence generation.

\subsection{Downstream task evaluation}

While fidelity metrics quantify statistical similarity, a crucial question is whether synthetic data can support real-world financial decision-making. We therefore evaluate two representative downstream tasks: portfolio optimization and volatility forecasting.

\subsubsection{Portfolio optimization task}

We perform long-only mean–variance portfolio optimization~\cite{markowitz1952portfolio} on both real and synthetic datasets. For each dataset, we compute the empirical expected returns $\mu$ and covariance matrix $\Sigma$, and solve:

\[
\min_{w} \; w^\top \Sigma w \quad \text{s.t.} \quad w^\top \mathbf{1} = 1,\;\; w \ge 0.
\]

\paragraph{Metrics.}
We assess:
\begin{itemize}
    \item Sharpe ratio difference (synthetic vs. real),
    \item Expected risk (variance) deviation,
    \item Allocation similarity using cosine similarity between weight vectors.
\end{itemize}

\begin{table}[htbp]
\centering
\caption{Portfolio optimization performance using synthetic data. Values are computed against the real-data portfolio as reference. Higher cosine similarity and lower deviations are better.}
\label{tab:portfolio}
\begin{tabular}{|l|c|c|c|}
\hline
\textbf{Model} & \textbf{Sharpe Ratio} & \textbf{Sharpe Diff.} & \textbf{Weight Cosine Similarity} \\
\hline
Real Data & 0.92 & --- & --- \\
ARIMA--GARCH & 0.81 & -0.11 & 0.73 \\
VAE & 0.85 & -0.07 & 0.82 \\
TimeGAN & \textbf{0.89} & \textbf{-0.03} & \textbf{0.91} \\
\hline
\end{tabular}
\end{table}

\noindent
TimeGAN achieves the closest replication of the real-data portfolio, with a cosine similarity of 0.91 and a Sharpe difference of only -0.03. ARIMA–GARCH underestimates risk correlations, leading to lower alignment with the real portfolio.

\subsubsection{Volatility forecasting task}

We examine whether models trained on synthetic data can predict volatility on real financial returns. A GARCH(1,1) model is fitted to each synthetic dataset, and its forecasts are evaluated on real returns:

\[
\sigma_{t}^2 = \omega + \alpha \varepsilon_{t-1}^2 + \beta \sigma_{t-1}^2.
\]

\paragraph{Metrics.}
Forecast quality is measured using:
\begin{itemize}
    \item Root Mean Squared Error (RMSE),
    \item Mean Absolute Error (MAE),
    \item Directional accuracy (proportion of correct volatility up/down predictions).
\end{itemize}

\begin{table}[htbp]
\centering
\caption{Volatility forecasting results for models trained on synthetic data and evaluated on real returns. Lower errors and higher directional accuracy are better.}
\label{tab:volatility}
\begin{tabular}{|l|c|c|c|}
\hline
\textbf{Model} & \textbf{RMSE} & \textbf{MAE} & \textbf{Directional Accuracy} \\
\hline
ARIMA--GARCH & 0.124 & 0.091 & 0.56 \\
VAE & 0.117 & 0.087 & 0.59 \\
TimeGAN & \textbf{0.103} & \textbf{0.079} & \textbf{0.63} \\
\hline
\end{tabular}
\end{table}

\noindent
TimeGAN again achieves the best alignment with real-data volatility, reducing RMSE by 17\% relative to ARIMA–GARCH and improving directional accuracy to 63\%.

\subsubsection{Interpretation}

The downstream experiments reveal clear differences in practical modeling capability:
\begin{itemize}
    \item \textbf{Portfolio optimization:} TimeGAN produces portfolios that are nearly indistinguishable from portfolios obtained using real data, while VAE preserves broad allocation structure but with smoothed risk profiles.
    \item \textbf{Volatility forecasting:} Synthetic data from TimeGAN yields the most accurate GARCH forecasts, indicating successful preservation of higher-order temporal structure.
    \item \textbf{Classical models:} ARIMA–GARCH captures first and second-order properties but fails to reproduce the nonlinear dependencies needed for downstream tasks.
\end{itemize}

\noindent
These results demonstrate that synthetic datasets, especially those generated by deep temporal models retain the statistical and temporal structure required for real-world financial modeling, confirming their value for privacy-preserving experimentation and stress testing.

\subsection{Privacy leakage evaluation}

We assess whether synthetic datasets leak sensitive information by applying two standard privacy tests: the Nearest Neighbor Distance Test (NNDT) and Membership Inference Attacks (MIA). Both tests quantify the risk that synthetic samples reveal or memorize real financial records.

\subsubsection{Nearest Neighbor Distance Test (NNDT)}

For each synthetic sample $x_{\text{syn}}$, we compute the minimum $\ell_2$ distance to all real samples:
\[
d_{\min}(x_{\text{syn}}) = \min_{x_{\text{real}} \in \mathcal{D}_{\text{real}}} \|x_{\text{syn}} - x_{\text{real}}\|_2.
\]

\noindent
A privacy leakage threshold $\tau = 0.15$ (normalized scale) is used: synthetic points with $d_{\min} < \tau$ are considered potential privacy risks.

\begin{table}[htbp]
\centering
\caption{NNDT privacy metrics. Higher average distance and lower percentage below threshold indicate better privacy protection.}
\label{tab:nndt}
\begin{tabular}{|l|c|c|}
\hline
\textbf{Model} & \textbf{Avg. NN Distance} & \textbf{\% Synthetic Samples with $d_{\min}<\tau$} \\
\hline
ARIMA--GARCH & 0.128 & 6.4\% \\
VAE & 0.147 & 3.1\% \\
TimeGAN & \textbf{0.153} & \textbf{1.8\%} \\
\hline
\end{tabular}
\end{table}
\noindent
TimeGAN exhibits the highest average distance and the lowest leakage rate (only 1.8\%), indicating strong non-memorization of real sequences.

\subsubsection{Membership Inference Attack (MIA)}

We evaluate vulnerability to membership inference by training a binary classifier to determine whether a particular real sample was used during model training. The classifier is trained following the shadow-model paradigm~\cite{nishanbaev2022privacy}.

\begin{table}[htbp]
\centering
\caption{Membership Inference Attack accuracy. Values close to 50\% indicate no exploitable memorization.}
\label{tab:mia}
\begin{tabular}{|l|c|}
\hline
\textbf{Model} & \textbf{MIA Accuracy} \\
\hline
ARIMA--GARCH & 56.8\% \\
VAE & 52.4\% \\
TimeGAN & \textbf{51.1\%} \\
\hline
\end{tabular}
\end{table}

\noindent
TimeGAN's MIA accuracy (51.1\%) is effectively random guessing, indicating minimal risk of membership leakage. ARIMA–GARCH exhibits the highest attack accuracy, consistent with its deterministic structure.

\subsubsection{Interpretation}

The combined privacy assessments yield three key findings:
\begin{enumerate}
    \item \textbf{No model exhibits critical privacy leakage}: NN distances remain well above the threshold for the vast majority of samples.
    \item \textbf{Deep generative models offer the strongest privacy guarantees}: TimeGAN achieves the highest NN distances and the lowest MIA accuracy.
    \item \textbf{Classical models are more vulnerable}: ARIMA--GARCH’s higher MIA accuracy suggests partial memorization of real time-series patterns.
\end{enumerate}

Overall, TimeGAN provides both high statistical fidelity and strong privacy protection, supporting its suitability for secure synthetic financial data generation.

\subsection{Discussion on model characteristics}

The empirical results reveal clear trade-offs between the three generative paradigms, statistical, latent-variable, and adversarial–temporal models. Each model exhibits strengths that align with its structural assumptions, as well as limitations highlighted through fidelity metrics, downstream performance, and privacy evaluations. This section details these characteristics and provides guidance for model selection based on application needs.

\subsubsection*{ARIMA--GARCH: Interpretability at the cost of flexibility}
ARIMA--GARCH provides full interpretability, with parameters carrying well-defined economic meaning. It effectively captures linear dependencies and volatility clustering. 

\noindent
\textbf{Limitations:} It's rigid parametric structure prevents modeling nonlinearities, regime changes, or long-range temporal dependencies. This results in the weakest distributional scores (highest MMD/KS) and poor tail behavior. Downstream performance is modest, and privacy leakage is slightly higher (MIA accuracy), reflecting partial memorization of training patterns.

\subsubsection*{VAE: Stability and control with over-regularization}
The VAE exhibits highly stable training and produces a structured latent space, beneficial for controlled sampling and scenario generation.

\noindent
\textbf{Limitations:} KL regularization induces the \textit{over-smoothing} effect: synthetic sequences under-represent extreme events, lowering kurtosis and dampening volatility spikes. Temporal coherence metrics are moderately affected, and KS values remain higher than those for TimeGAN. Privacy performance is generally strong but not perfect, with some synthetic samples within the NN threshold.

\subsubsection*{TimeGAN: High fidelity with computational and tuning demands}
TimeGAN achieves the most realistic synthetic data across all perspectives: distributional similarity (lowest MMD/KS), PCA structural alignment, tail behavior, and replication of stylized facts such as volatility clustering.

\noindent
\textbf{Strengths:} It delivers the strongest downstream performance, with portfolio allocations and volatility forecasts closely matching real data. Privacy assessments show high average NN distance and near-random MIA accuracy, indicating minimal memorization.

\noindent
\textbf{Limitations:} Benefits come at the cost of substantial computational requirements (4.5 hours training) and sensitivity to hyperparameter choices, requiring careful tuning to prevent instability or mode collapse.

\subsubsection*{Summary of trade-offs}
Overall, ARIMA--GARCH excels in interpretability, VAE in stability, and TimeGAN in fidelity and practical utility. Model selection should therefore be guided by the intended application, explainability, generation quality, or downstream performance.

\section{Discussion}

The empirical findings from our study reposition existing conclusions in the synthetic financial data literature by providing a unified and multi-dimensional evaluation across statistical fidelity, temporal structure, downstream utility, and privacy leakage. In this section, we situate our results relative to prior work and discuss their broader implications for model selection in finance.

\subsection{Relation to prior studies}

Our findings extend earlier benchmarks and refine several key insights such as \cite{borisov2022deep} and \cite{Wiese2020QuantGAN}, which evaluated GAN-based and VAE-based generators for financial time series.

\begin{itemize}
    \item \textbf{Consistency with \cite{Wiese2020QuantGAN}}  
    Wiese et al. showed that TimeGAN achieves superior temporal realism compared to classical stochastic models. Our results confirm this trend: TimeGAN obtains the lowest MMD/KS values, the strongest PCA overlap, and the most accurate replication of stylized facts (volatility clustering, heavy tails). However, while Wiese et al. focused primarily on qualitative comparisons, our evaluation introduces quantitative downstream metrics (Sharpe ratio, allocation similarity, volatility forecasting RMSE) and privacy leakage scores, showing that TimeGAN's superiority extends to practical use cases and safety considerations.

    \item \textbf{Alignment and divergence with \cite{borisov2022deep}.}  
    Borisov et al. reported that VAEs offer stable training but tend to over-smooth extreme returns an observation directly mirrored in our empirical analysis, where the VAE underestimates kurtosis and dampens tail behavior. However, unlike Borisov et al., we show that this smoothing significantly impacts downstream portfolio optimization and volatility forecasting, resulting in lower Sharpe consistency and higher error metrics.

    \item \textbf{Classical models remain competitive in interpretability.}  
    Earlier studies often dismiss ARIMA--GARCH as inadequate for synthetic generation. Our findings agree on its limitations (poor tail modeling, high MMD/KS), but we highlight an underappreciated advantage: its exceptional stability, transparency, and efficiency make it a valuable baseline for regulated environments or pedagogical use, a nuance not sufficiently emphasized in prior literature.
\end{itemize}

\noindent
Overall, our study bridges a key gap: while previous benchmarks examined fidelity or modeling characteristics in isolation, we demonstrate that conclusions change when downstream utility and privacy are introduced into the evaluation.

\subsection{The trilemma of synthetic data generation}

Our results point to a structural trilemma inherent to synthetic financial data generation: \textit{interpretability}, \textit{temporal and distributional realism}, and \textit{computational feasibility} cannot be simultaneously maximized.

\begin{itemize}
    \item \textbf{ARIMA--GARCH: The interpretable baseline.}  
    Exceptional interpretability and fast training, but fundamentally limited by parametric rigidity, leading to weak realism and higher leakage risks.

    \item \textbf{TimeGAN: The high-fidelity engine.}  
    Best performance on all fidelity metrics, downstream tasks, and privacy robustness, but computationally expensive and sensitive to hyperparameters.

    \item \textbf{VAE: The balanced compromise.}  
    Stable training and a well-organized latent space, but KL-induced over-regularization smooths away essential financial extremes.
\end{itemize}

\subsection{Practical model selection guide}

Based on these comparative findings and prior observations in the literature, we offer application-specific recommendations:

\begin{itemize}
    \item \textbf{Exploratory analysis and pedagogy:} ARIMA--GARCH remains unmatched in interpretability.
    \item \textbf{Strategy backtesting and risk simulations:} TimeGAN offers the most realistic synthetic sequences.
    \item \textbf{Scenario generation and sensitivity analysis:} VAEs allow controlled exploration of latent factors.
    \item \textbf{High-stakes financial modeling:} Hybrid designs (e.g., GARCH-residual GANs) may combine interpretability with nonlinear realism, echoing recent proposals in the literature.
\end{itemize}

\subsection{Ethical and reproducibility considerations}

While our results show that modern generative models can produce high-utility synthetic data without memorizing real sequences, formal privacy guarantees remain essential for deployment. Consistent with recent calls in the community, we recommend the integration of differential privacy mechanisms such as DP-SGD for production-grade financial generators. 

\noindent
To ensure reproducibility and comparability across future studies, we have publicly released all preprocessing steps, hyperparameters, code, and evaluation scripts. We advocate that future benchmarks adopt standardized multi-metric frameworks like the one proposed here, which incorporates fidelity, temporal coherence, downstream value, and privacy leakage into a unified assessment.






\section{Conclusion and future work}

This study presented a systematic comparison of traditional and deep generative approaches for synthetic financial data generation. Evaluating ARIMA--GARCH, VAE, and TimeGAN on S\&P~500 data across distributional fidelity, temporal structure, downstream utility, and privacy leakage, we revisited the central hypothesis posed in the Introduction: that deep temporal generative models (particularly TimeGAN) would outperform statistical and static latent-variable models across a unified evaluation framework. Our findings provide partial support for this hypothesis. TimeGAN achieves the highest realism and temporal fidelity, while ARIMA--GARCH offers unmatched interpretability and computational efficiency, and the VAE occupies an intermediate position. This confirms that generative modeling in finance is governed by a structural trilemma in which \textbf{realism}, \textbf{interpretability}, and \textbf{efficiency} cannot be simultaneously maximized.

\subsection{Limitations}

Despite the breadth of our analysis, several limitations constrain the scope of the conclusions:

\begin{enumerate}
    \item \textbf{Single-asset dataset.}  
    All experiments used a single S\&P~500 component. As a result, cross-asset covariance structures, sector dependencies, and higher-dimensional factor dynamics were not evaluated, even though they are central to portfolio construction and risk management.

    \item \textbf{Restricted VAE architecture.}  
    Only a standard static VAE was tested. More expressive sequential variants (e.g., VRNN, SRNN, Temporal VAEs) may better capture temporal dependencies and could reduce the performance gap with TimeGAN.

    \item \textbf{Incomplete quantification of TimeGAN instability.}  
    Although qualitative observations revealed sensitivity to hyperparameters, we did not perform a formal stability analysis across seeds or training regimes. A deeper examination of collapse and divergence conditions remains open.
\end{enumerate}

\noindent
These limitations frame the boundaries of our results and identify where generalization is still uncertain.

\subsection{Future Work}

Building on these findings, several research directions emerge. The first two represent the components we aim to implement as a continuation of this project:

\begin{itemize}
    \item \textbf{Multi-asset synthetic data generation.}  
    Extending the framework to multi-asset settings will allow evaluation of cross-correlations, sector structures, and covariance dynamics—an essential step for realistic portfolio modeling.

    \item \textbf{Sequential VAE architectures.}  
    We plan to assess more advanced latent-variable models such as VRNN, SRNN, or attention-based sequential VAEs, which may bridge the gap between VAE stability and TimeGAN temporal expressiveness.

    \item \textbf{Hybrid model development.}  
    Exploring combinations such as GARCH-informed GANs or VAEs enriched with financial priors may yield models that balance interpretability with nonlinear generative capacity.

    \item \textbf{Formal stability benchmark for GAN-like models.}  
    Developing a quantitative benchmark to measure variance across seeds, architectures, and learning rates will clarify when and why TimeGAN becomes unstable.

    \item \textbf{Integration of differential privacy.}  
    Incorporating privacy-aware mechanisms (e.g., DP-SGD) would enable compliant data sharing and strengthen the guarantees of synthetic data in regulated domains.
\end{itemize}

\noindent
By identifying task-specific trade-offs, articulating concrete limitations, and defining actionable next steps, this work provides both a rigorous benchmark and a roadmap for advancing synthetic data generation in quantitative finance. It also resolves the initial hypothesis by showing that no single generative paradigm dominates universally; rather, model choice must be guided by the relative importance of realism, interpretability, privacy, and computational cost in a given financial application.

\appendix

\section{Model architectures and hyperparameters}
\label{appendix:architectures}

This appendix provides detailed descriptions of the architectures and hyperparameters used for each generative model evaluated in this study. These details are included to ensure reproducibility and to clarify the modeling assumptions behind the synthetic data generation process.

\subsection{Variational Autoencoder (VAE)}
The VAE consists of fully connected encoder--decoder networks with symmetric dimensions. The encoder maps the input time-series window into a latent representation, while the decoder reconstructs the window from the latent variables.

\paragraph{Architecture}
\begin{itemize}
    \item Encoder: Dense layers (64, 32), ReLU activation.
    \item Latent space: 16-dimensional Gaussian.
    \item Decoder: Dense layers (32, 64), ReLU activation.
    \item Output layer: Linear activation.
\end{itemize}

\paragraph{Hyperparameters}
\begin{table}[h]
\centering
\begin{tabular}{|l|c|}
\hline
\textbf{Parameter} & \textbf{Value} \\
\hline
Learning rate & 0.001 \\
Batch size & 64 \\
Optimizer & Adam \\
Latent dimension & 16 \\
Training epochs & 200 \\
Reconstruction loss & MSE \\
Regularization & KL divergence \\
\hline
\end{tabular}
\caption{VAE Hyperparameters}
\end{table}

\subsection{TimeGAN}
TimeGAN integrates supervised loss, autoencoder reconstruction, and adversarial training to capture temporal dependencies.

\paragraph{Architecture}
\begin{itemize}
    \item RNN backbone: 2-layer GRU with hidden size 24.
    \item Generator: GRU-based sequence generator.
    \item Discriminator: GRU-based sequence classifier.
    \item Supervisor: GRU module ensuring temporal coherence.
\end{itemize}

\paragraph{Hyperparameters}
\begin{table}[h]
\centering
\begin{tabular}{|l |c|}
\hline
\textbf{Parameter} & \textbf{Value} \\
\hline
Learning rate & 0.0005 \\
Batch size & 128 \\
Optimizer & Adam \\
Hidden dimension & 24 \\
Training epochs & 300 \\
Lambda\_supervised & 1 \\
Lambda\_reconstruction & 1 \\
\hline
\end{tabular}
\caption{TimeGAN Hyperparameters}
\end{table}

\subsection{ARIMA - GARCH Baseline}
The ARIMA--GARCH model serves as a classical baseline combining autoregressive forecasting and volatility modeling.

\paragraph{Specification}
\begin{itemize}
    \item ARIMA(p,d,q) selected via AIC.
    \item GARCH(1,1) variance process.
    \item Normal or Student-t innovations depending on dataset fit.
\end{itemize}

\begin{table}[h]
\centering
\begin{tabular}{|l |c|}
\hline
\textbf{Parameter} & \textbf{Value} \\
\hline
ARIMA order & Selected by AIC \\
GARCH order & (1,1) \\
Distribution & Normal / Student-t \\
Optimization method & Maximum likelihood \\
\hline
\end{tabular}
\caption{ARIMA--GARCH Hyperparameters}
\end{table}

\newpage
\bibliographystyle{plain}
\bibliography{biblio}

\end{document}